\documentclass{article}
\usepackage{spconf, amsmath, graphicx}
\usepackage{color, algorithm, algpseudocode, multirow, amssymb, array, float}

\usepackage{booktabs} 
\usepackage{multicol}		
\usepackage{multirow} 		
\usepackage{bbding}			

\title{SAM-OCTA: A Fine-Tuning Strategy for Applying Foundation Model to OCTA Image Segmentation Tasks}
\twoauthors
 {Chengliang Wang, Xinrun Chen, Haojian Ning}
	{Chongqing University\\
	College of Computer Science\\
	Chongqing, China}
 {Shiying Li}
	{Xiang’an Hospital of Xiamen University\\
	Department of Ophthalmology \\
	Xiamen, China}
\begin{document}
%
\maketitle
\begin{abstract}
In the analysis of optical coherence tomography angiography (OCTA) images, the operation of segmenting specific targets is necessary. Existing methods typically train on supervised datasets with limited samples (approximately a few hundred), which can lead to overfitting. To address this, the low-rank adaptation technique is adopted for foundation model fine-tuning and proposed corresponding prompt point generation strategies to process various segmentation tasks on OCTA datasets. This method is named SAM-OCTA and has been experimented on the publicly available OCTA-500 dataset. While achieving state-of-the-art performance metrics, this method accomplishes local vessel segmentation as well as effective artery-vein segmentation, which was not well-solved in previous works. The code is available at: https://github.com/ShellRedia/SAM-OCTA.
\end{abstract}

\vspace{-4pt}
\section{Introduction}
\label{sec:Introduction}
\vspace{-2pt}

Optical coherence tomography angiography (OCTA) is an innovative and non-invasive imaging technique that enables the visualization of retinal microvasculature with high resolution and without needing dye injection \cite{wang2021deep}. It is a valuable tool for disease staging and preclinical diagnosis \cite{liang2021foveal}.

Certain specific retinal structures, such as retinal vessels (RV) and the avascular zone (FAZ) of the macula, usually need to be segmented from the raw data of OCTA for further analysis \cite{liang2021foveal, li2022rps}. Researchers have been actively exploring deep learning-based methods for image quality assessment and segmentation to address these challenges and enhance the accuracy and efficiency of OCTA image analysis. Most deep learning segmentation methods related to OCTA are based on self-designed neural networks and modules. This requires training the model from scratch, which can lead to overfitting issues. Foundational models, trained on large-scale data, can be applied to various scenarios \cite{zhang2023comprehensive}.

Segment Anything Model (SAM) was introduced as a foundational model for addressing natural image tasks. This benchmark model demonstrated, for the first time, the promising wide applicability to various image segmentation tasks without the need for prior re-training \cite{kirillov2023segment}. However, medical images differ significantly from natural images in terms of quality, noise, resolution, and other factors, which can affect SAM's segmentation performance. Thus, further research and optimization efforts are required to fully harness the potential of SAM in medical image segmentation \cite{zhang2023segment}. 

We find that adopting a fine-tuning approach to SAM and introducing prompt information can enhance and guide the model's segmentation, aiming to improve some complex OCTA segmentation cases. We call our method as SAM-OCTA and summarize the contributions as follows: 

(1) Applying Low-Rank Adaptation (LoRA) technology for fine-tuning the SAM model enables it to perform effective segmentation of specific targets within OCTA images.

(2) A strategy for generating prompt points has been proposed, which enhances the segmentation performance of FAZ and artery-vein tasks within OCTA samples.

\vspace{-5pt}
\section{Related Work}
\label{sec:RelatedWork}

\vspace{-5pt}
\subsection{OCTA Segmentation Models}
\vspace{-2pt}
As a typical architecture for deep image processing, the vision transformer (ViT) is frequently used for segmentation tasks in OCTA \cite{dosovitskiy2020image}. In OCTA images, the distribution of RV is extensive, and it requires the models to effectively utilize the global information in the images. The TCU-Net, OCT2Former, and StruNet methods have improved ViT, achieving continuous RV segmentation and addressing issues such as vessel discontinuities or missing segments \cite{shi2023tcu, tan2023oct2former, ma2023strunet}. Other methods, from the perspectives of efficiency, denoising, and the utilization of three-dimensional data, have designed a series of techniques and strategies, achieving promising segmentation results on OCTA datasets~\cite{liang2021foveal, wang2023db, li2020image, zhu2022ovs, ma2022retinal}. The above-mentioned methods have demonstrated that existing deep networks are capable of achieving precise segmentation of RV and FAZ. 

\vspace{-4pt}
\subsection{SAM and Related Fine-tuning Approaches}
\vspace{-2pt}
The SAM is a foundational vision model for general image segmentation. With the ability to segment diverse objects, parts, and visual structures in various scenarios, SAM takes prompts in the form of points, bounding boxes, or coarse masks as input. Its remarkable zero-shot segmentation capabilities enable its easy transfer to numerous applications through simple prompting \cite{kirillov2023segment}. Although SAM has established an efficient data engine for model training, there are relatively few cases collected for medical applications or other rare image scenarios. Therefore, some fine-tuning methods have been applied to SAM to improve its performance in certain segmentation failure cases \cite{zhang2023customized, wu2023medical}. The common characteristic of these fine-tuning methods is that they introduce additional network layers on top of the pre-trained SAM. By adding a small number of trainable parameters, fine-tuning becomes feasible through training on the new dataset. The advantage of fine-tuning methods lies in their ability to preserve SAM's strong zero-shot capabilities and flexibility.

\vspace{-2pt}
\section{Method}
\label{sec:Method}
\vspace{-2pt}

\begin{figure}
  \centering
  \includegraphics[width=1\linewidth]{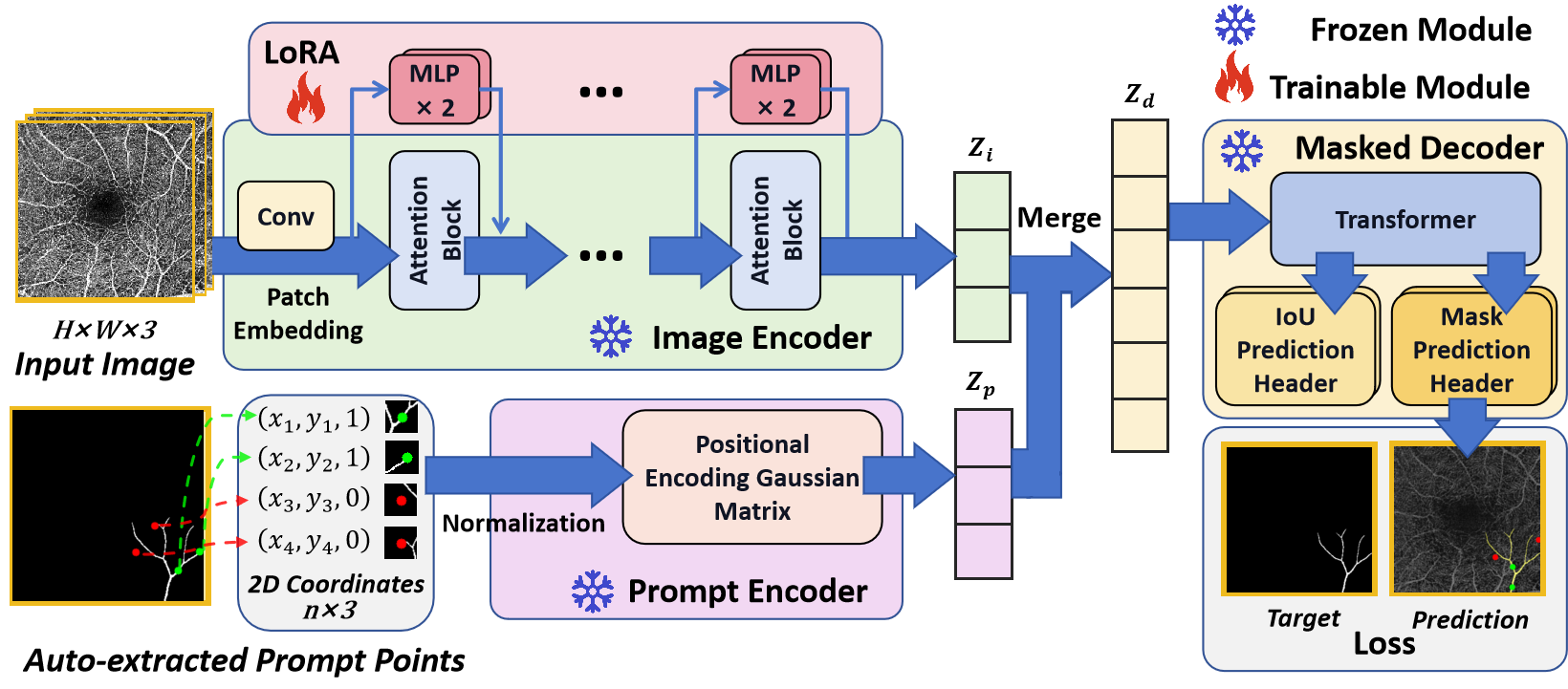}
  \caption{Schematic diagram illustrating the fine-tuning of SAM using OCTA samples.}
  \label{Architecture}
\end{figure}

In this paper, we fine-tuned the pre-trained SAM using OCTA datasets and corresponding annotations. The process is shown in Figure  \ref{Architecture}. SAM consists of three parts: an image encoder, a flexible prompt encoder, and a fast mask decoder \cite{kirillov2023segment}. 

\vspace{-2pt}
\subsection{Fine-tuning of Image Encoder}
\vspace{-2pt}

The image encoder utilizes a ViT pre-trained with the masked auto-encode method. The ViT model comes in three variants: vit-b, vit-l, and vit-h which can only process fixed-size inputs (e.g. $1024 * 1024 * 3$). To support input images of different resolutions, scaling and padding operations are employed. In this study, we used the image encoder from the "vit-h" model for the fine-tuning process.

\begin{figure}
  \centering
  \includegraphics[width=1\linewidth]{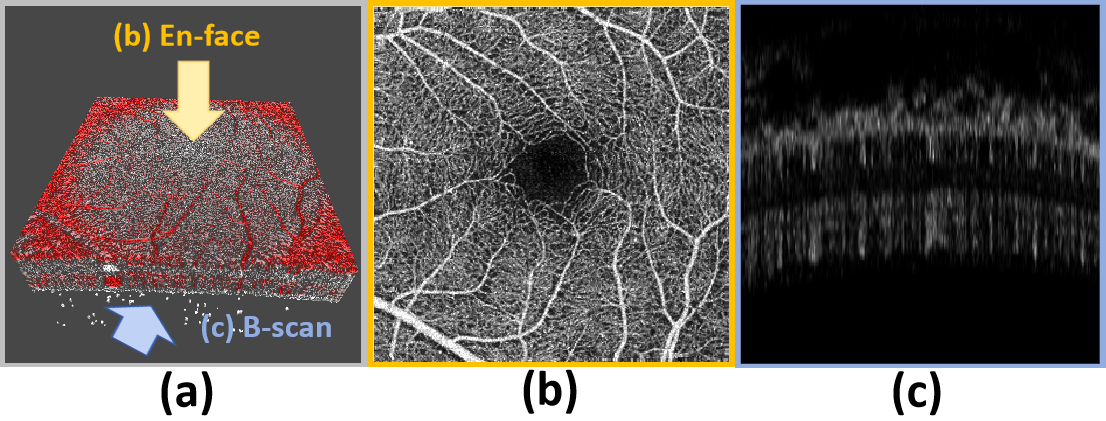}
  \caption{OCTA Structural Diagram. (a) Three-dimensional volume rendering with arrows indicating different projection directions. (b) En-face projection. (c) B-scan projection.}
  \label{Fig:OCTA-Structure}
\end{figure}

As shown in Figure \ref{Fig:OCTA-Structure}, OCTA data is inherently in 3D format, but most datasets provide en-face 2D projection forms. En-face projection is obtained through layer-wise segmentation based on vascular anatomical structures. As SAM requires three-channel images as input, in this work, we stack projection layers in different depths of OCTA images to adapt to this input format. The benefit of this approach is that it preserves the vascular structure information in the OCTA images while fully utilizing SAM's feature-extracting capabilities. Fine-tuning aims to retain SAM's powerful image-understanding capabilities while enhancing its performance on OCTA. The approach used in this paper involves utilizing the LoRA technique \cite{hu2021lora}, which introduces additional linear network layers in each transformer block of the image encoder, similar in form to a ResNet block. During the training process, the weights of the SAM are frozen, and only the newly introduced parameters are updated.

\vspace{-2pt}
\subsection{Prompt Points Generation Strategy}

The prompt encoder is divided into two types of prompts: sparse prompts (points, boxes, text) and dense prompts (masks). In our work, we chose points as the prompt for OCTA segmentation. For each sample's prompt points input, assuming there are $n$ points in the prompt point input, it can be represented as ${(x_1, y_1, 1), (x_2, y_2, 1), ...,(x_n, y_n, 0)}$, where $x$ and $y$ denote the coordinates of prompt points in the image. The values "1" and "0" indicate positive (foreground) and negative (background) points, respectively. 
The prompt encoder of SAM will perform embedding on this input, and due to its pre-training, it can appropriately integrate with the information from the input image. 

\begin{figure}
  \centering
  \includegraphics[width=1\linewidth]{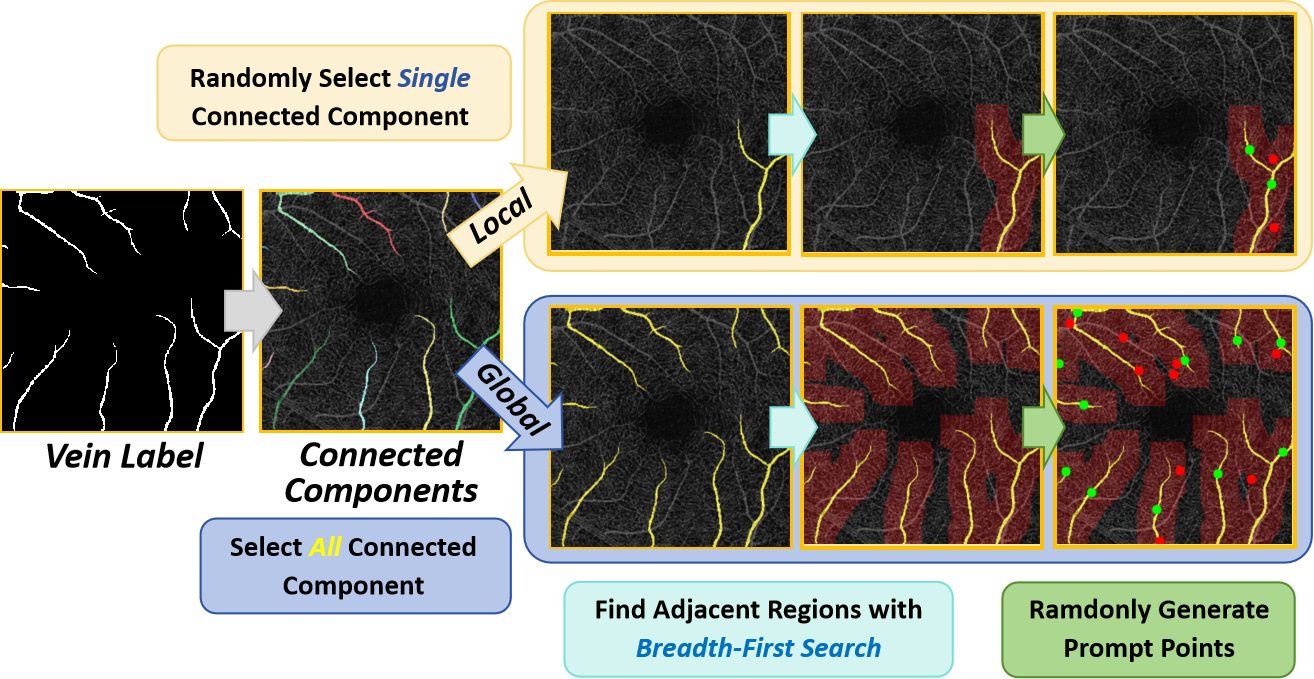}
  \caption{Illustration of Prompt Points Generation. Green and red points represent positive and negative points, respectively.}
  \label{Fig:PPG}
\end{figure}

The prompt points generation strategy has two types: the global mode and the local mode. The global mode is applied to all OCTA segmentation tasks, while the local mode is specific to artery/vein segmentation. The study of segmenting individual vessels, as a local segmentation task, has not been attempted in previous works. By the prompt encoder, more accurate regional vessel segmentation can be achieved in OCTA datasets. For this task, the first step is to identify and label all connected components in the segmentation masks with unique identifiers. Due to the weak connectivity at the endpoints of some vessels in OCTA labels, we adopt the eight-connectivity criterion. Then positive points are randomly selected within each connected component. Due to varying numbers of vessels in different samples, to standardize their data format, negative points are added from the background adjacent to the connected components. The prompt points generation process can be described as Figure \ref{Fig:PPG}.

\vspace{-2pt}
\subsection{Mask Decoder}
\vspace{-2pt}

The role of the mask decoder is to efficiently map the image embeddings, prompt embeddings, and output tokens to a segmentation mask. A modified version of the transformer decoder block is employed, followed by a dynamic mask prediction head. For an image input and corresponding prompt input, the mask decoder outputs multiple segmentation masks to represent objects at different semantic levels. In this work, the loss function (in the fine-tuning process) is computed based on the segmentation output with the highest confidence.


\begin{table*}[t]
\centering
\caption{RV and FAZ Segmentation Results on OCTA-500 (underscores indicate the top two highest values).}
\label{GlobalResult}
\begin{tabular}{ccccccccc}
\toprule
Label & \multicolumn{4}{c}{RV} & \multicolumn{4}{c}{FAZ} \\
\midrule
\multirow{2}{*}{Method} & \multicolumn{2}{c}{OCTA-500(3M)} & \multicolumn{2}{c}{OCTA-500(6M)} & \multicolumn{2}{c}{OCTA-500(3M)} & \multicolumn{2}{c}{OCTA-500(6M)} \\
\cmidrule(lr){2-3} \cmidrule(lr){4-5} \cmidrule(lr){6-7} \cmidrule(lr){8-9}
& Dice ↑ & Jaccard ↑ & Dice ↑ & Jaccard ↑ & Dice ↑ & Jaccard ↑ & Dice ↑ & Jaccard ↑ \\
\midrule
    U-Net (2015) & 0.9068 & 0.8301 & 0.8876 & 0.7987 & 0.9747 & 0.9585 & 0.8770 & 0.8124 \\
      IPN (2020) & 0.9062 & 0.8325 & 0.8864 & 0.7973 & 0.9505 & 0.9091 & 0.8802 & 0.7980 \\
  IPN V2+ (2021) & \underline{0.9274} & \underline{0.8667} & \underline{0.8941} & \underline{0.8095} & 0.9755 & 0.9532 & \underline{0.9084} & 0.8423 \\
    FARGO (2021) & 0.9112 & 0.8374 & 0.8798 & 0.7864 & 0.9785 & 0.9587 & 0.8930 & 0.8355 \\
Joint-Seg (2022) & 0.9113 & 0.8378 & \underline{0.8972} & \underline{0.8117} & \underline{0.9843} & \underline{0.9693} & 0.9051 & \underline{0.8424} \\
\midrule
SAM-OCTA (ours) & \underline{\textbf{0.9199}} & \underline{\textbf{0.8520}} & 0.8869 & 0.7975 & \underline{\textbf{0.9838}} & \underline{\textbf{0.9692}} & \underline{\textbf{0.9073}} & \underline{\textbf{0.8473}} \\
\bottomrule
\end{tabular}
\end{table*}

\vspace{-2pt}
\section{Experiments}
\vspace{-2pt}

\subsection{Datasets and Preprocessing}

The publicly available dataset used in this paper is OCTA-500~\cite{li2020ipn}. The OCTA-500 dataset contains 500 samples, classified based on the field of view (FoV): $3mm * 3mm$ (3M) and $6mm * 6mm$ (6M). The corresponding image resolutions are $304 * 304$ and $400 * 400$, with 200 and 300 samples respectively. The OCTA-500 dataset provides annotations for RV, FAZ, capillary, artery, and vein. The adopted data augmentation tool is Albumentations \cite{info11020125}. The data augmentation strategies include horizontal flipping, brightness and contrast adjustment, and random slight rotation. 

\subsection{Experimental Settings}

The SAM is deployed on A100 graphic cards with 80 GB memory. The 10-fold cross-validation is adopted to evaluate the training results. 
The optimizer used is AdamW, and the learning rate adopts a warm-up strategy, starting from $10^{-5}$ and gradually increasing to $10^{-3}$.

The loss functions used for fine-tuning vary depending on the segmentation tasks. For FAZ and capillary, the Dice loss is employed. However, for RV, artery, and vein, the clDice loss is utilized which is more feasible for tubular segmentation \cite{shit2021cldice}. These two loss functions can be represented as: 
\begin{align}
L_{clDice} = 0.2 * L_{Dice} + 0.8 * L_{clDice}^\prime, \nonumber
\end{align} 
where $ L_{Dice} = 1 - \frac{2 * |\hat{Y} \cap Y|}{|\hat{Y}| + |Y|} $, 

\ \ \ \ \ $ L_{clDice}^\prime = 1 - 2 * \frac{T{prec}(\hat{Y_s}, Y) * T{sens}(Y_s, \hat{Y})}{T{prec}(\hat{Y_s}, Y) + T{sens}(Y_s, \hat{Y})} $, 

\ \ \ \ \ $Y$ → {\itshape the ground-truth}, 

\ \ \ \ \ $\hat{Y}$ → {\itshape the predicted value}, 

\ \ \ \ \ $Y_s, \hat{Y_s}$ → {\itshape soft{-}skeleton($Y$, $\hat{Y}$)}, and 

\ \ \ \ \ $T{prec}, T{sens}$ → {\itshape precision and sensitivity.}

\subsection{Results}

We conducted extensive experiments with various cases on the OCTA datasets. The segmentation results using metrics Dice, and Jaccard, which are calculated as follows:
\begin{align}
Dice(\hat{Y}, Y) = \frac{2 |\hat{Y} \cap Y|}{|\hat{Y}| + |Y|}, \ \ Jaccard(\hat{Y}, Y) = \frac{|\hat{Y} \cap Y|}{|\hat{Y} \cup Y|}. \nonumber
\end{align}



\subsubsection{Global Mode}

 We have experimented with various prompt point generation strategies, including the number of points and the generation area for negative points, and have selected the best metrics as the final results. RV and FAZ are common segmentation tasks in previous studies. Therefore, we will summarize the comparative results in Table \ref{GlobalResult}. The experimental data from previous methods are referenced from \cite{hu2022joint}. Our method's comprehensive performance reaches the state-of-the-art level.

\begin{figure}[h]
  \centering
  \includegraphics[width=0.8\linewidth]{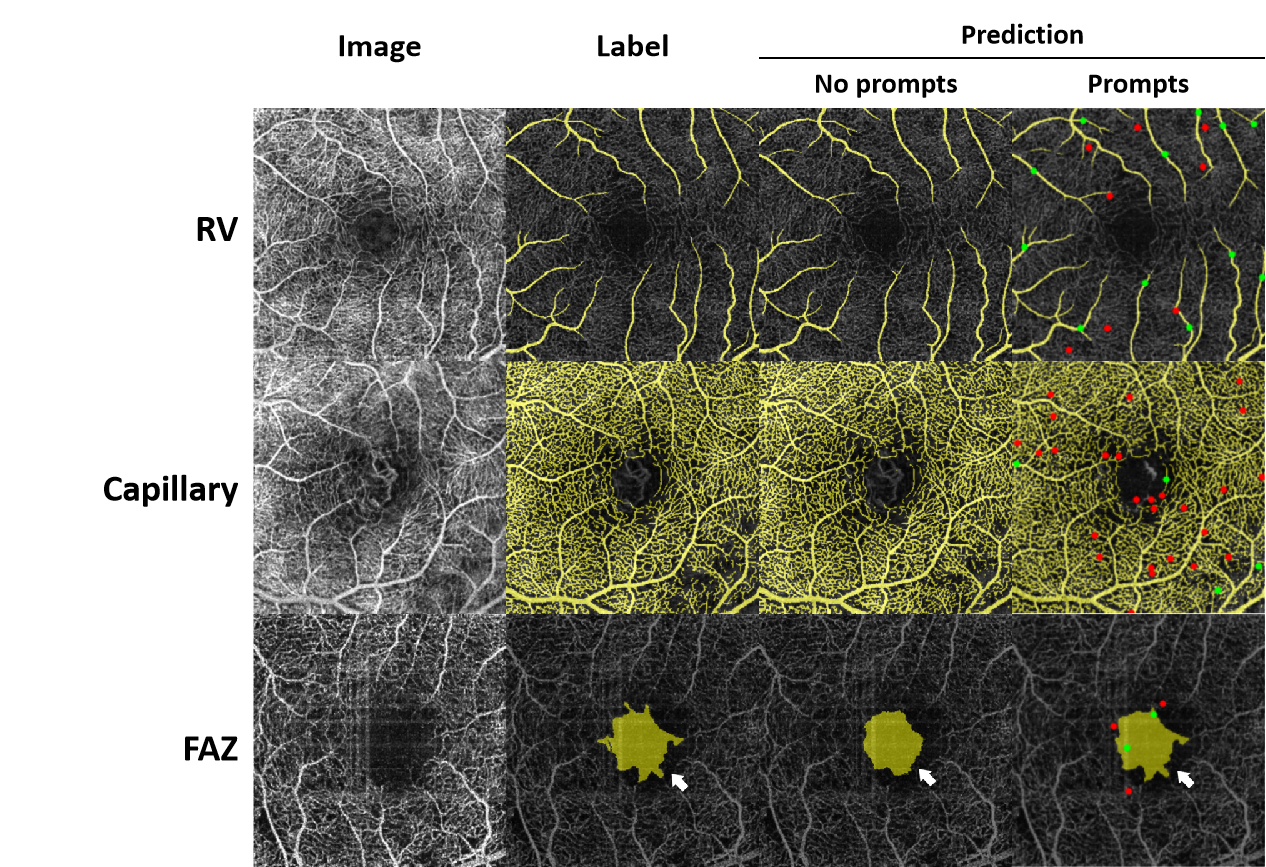}
  \caption{Segmentation results of SAM-OCTA in RV, capillary, and FAZ, with white arrows indicating areas of improvement with added prompt points.}
  \label{Fig:GlobalSeg}
\end{figure}

\begin{figure}[h]
  \centering
  \includegraphics[width=1\linewidth]{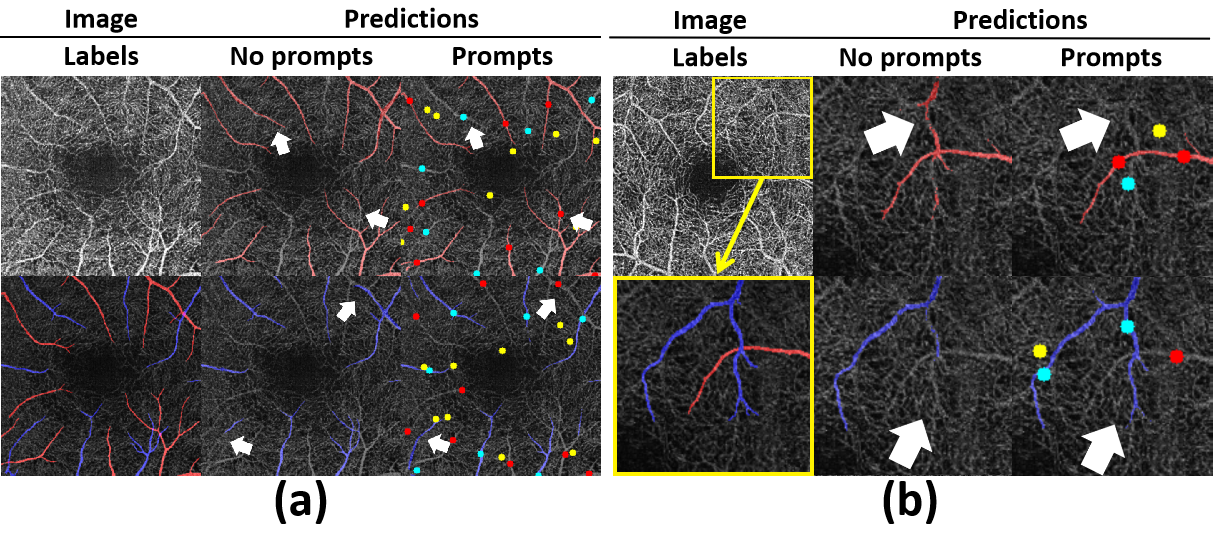}
  \caption{The performance of SAM-OCTA on artery and vein segmentation tasks. (a) Global mode; (b) Local mode. The red and blue vessels respectively represent arteries and veins. Red and cyan dots represent corresponding prompt points. Yellow dots represent negative background prompt points. For the artery segmentation, the red and cyan dots are respectively positive and negative points. For the vein, it should be the opposite. White arrows indicate areas of improvement with added prompt points.}
  \label{Fig:AVSeg}
\end{figure}

For segmentation tasks involving global vessels such as RV and capillary, the impact of prompt points is not significant. However, for FAZ, artery, and vein segmentation, prompt points lead to a noticeable improvement in segmentation performance. The segmentation results can be observed in Figures \ref{Fig:GlobalSeg}, \ref{Fig:AVSeg}, and Table \ref{PromptPointEffect}. It can be inferred that the effect of prompt point information is more pronounced within a local region. For the widely distributed vessels, importing more prompt points has a limited effect. However, the prompt points can help improve the boundary delineation of the FAZ.

\begin{table}[t]
\centering
\caption{The effect of prompt points on segmentation tasks.}
\label{PromptPointEffect}
\resizebox{\linewidth}{!}{
\begin{tabular}{cccccc}
\toprule
\multicolumn{2}{c}{FoV} & \multicolumn{2}{c}{OCTA-500(3M)} & \multicolumn{2}{c}{OCTA-500(6M)} \\
\cmidrule(lr){1-2} \cmidrule(lr){3-4} \cmidrule(lr){5-6}
\multicolumn{2}{c}{Prompts} & \textcolor{black}{{\scriptsize \XSolidBrush}} & {\scriptsize \CheckmarkBold} & \textcolor{black}{{\scriptsize \XSolidBrush}} & {\scriptsize \CheckmarkBold} \\ 
\midrule
&&\multicolumn{4}{c}{Global Mode} \\
\cmidrule(lr){3-6}
\multirow{2}{*}{RV} & Dice ↑ & 0.9165 & 0.9199 & 0.8865 & 0.8869 \\
& Jaccard ↑ & 0.8431 & 0.8520 & 0.7955 & 0.7975 \\
\multirow{2}{*}{FAZ} & Dice ↑ & 0.9545 & 0.9838 & 0.8787 & 0.9073 \\
& Jaccard ↑ & 0.9345 & 0.9692 & 0.7991 & 0.8473 \\
\multirow{2}{*}{Capillary} & Dice ↑ & 0.8813 & 0.8785 & 0.8337 & 0.8379 \\
& Jaccard ↑ & 0.7881 & 0.7837 & 0.7152 & 0.7213 \\
\multirow{2}{*}{Artery} & Dice ↑ & 0.8342 & 0.8747 & 0.8352 & 0.8602 \\
& Jaccard ↑ & 0.7528 & 0.7785 & 0.7325 & 0.7572 \\
\multirow{2}{*}{Vein} & Dice ↑ & 0.8409 & 0.8817 & 0.8263 & 0.8526 \\
& Jaccard ↑ & 0.7463 & 0.7897 & 0.7168 & 0.7474 \\
\cmidrule(lr){1-6}
&&\multicolumn{4}{c}{Local Mode} \\
\cmidrule(lr){3-6}
\multirow{2}{*}{Artery} & Dice ↑ & 0.7393 & 0.8707 & 0.6865 & 0.7922 \\
& Jaccard ↑ & 0.6339 & 0.7792 & 0.5699 & 0.6851 \\
\multirow{2}{*}{Vein} & Dice ↑ & 0.7742 & 0.8352 & 0.7053 & 0.8167 \\
& Jaccard ↑ & 0.6658 & 0.7267 & 0.5823 & 0.7014 \\

\bottomrule
\end{tabular}
}
\end{table}

\subsubsection{Local Mode}

The local mode primarily focuses on precisely segmenting vessels in local regions. The segmentation targets include the artery and vein. For each sample, two positive points are selected on the target vessels, and two negative points are selected on the adjacent region. 

Due to the morphological similarities between retinal arteries and veins, as well as the complexities introduced by factors such as age, gender, and disease conditions, deep learning methods often encounter segmentation disconnections or confusion in the artery-vein segmentation task \cite{xu2022av}. Without prompt points, the OCTA-SAM is prone to confusion when an artery and a vein are closed or overlapping. Table \ref{PromptPointEffect} reveals the substantial metrics improvement with prompts. From Figure \ref{Fig:AVSeg}, it can be seen that the introduced prompt points (especially the negative points on the vein when segmenting artery) assist the model in effectively distinguishing different types of vessels, thereby improving the artery-vein segmentation results. 

\vspace{-2pt}
\section{Conclusion}
\vspace{-2pt}

We propose a fine-tuning method for SAM for OCTA image segmentation and design prompt point generation strategies as global and local modes. It excels in both RV and FAZ tasks while also firstly exploring and achieving good results in the artery-vein segmentation task on the OCTA-500 dataset. This is expected to assist in the analysis and diagnosis of related diseases with varying impacts on arteries and veins.

\vspace{-2pt}
\section*{Acknowledgement}
\vspace{-2pt}

This work is supported by the Chongqing Technology Innovation $\And$ Application Development Key Project (cstc2020jscx; dxwtBX0055; cstb2022tiad-kpx0148).

\newpage

\bibliographystyle{IEEEbib}
\bibliography{main.bib}

\end{document}